\documentclass[10pt]{article} 
\usepackage[preprint]{tmlr}

\usepackage{amsmath,amsfonts,bm}

\def\eqref#1{equation~\ref{#1}}

\def\1{\bm{1}}

\DeclareMathAlphabet{\mathsfit}{\encodingdefault}{\sfdefault}{m}{sl}
\SetMathAlphabet{\mathsfit}{bold}{\encodingdefault}{\sfdefault}{bx}{n}

\usepackage{float}
\usepackage{graphicx}
\usepackage{tikz}
\usepackage{pgfplots}
\usepackage{amsmath}
\usepackage{hyperref}
\usepackage{url}
\usepackage{booktabs}
\usepackage{subcaption}

\title{ATA: A Neuro-Symbolic Approach to Implement \\ Autonomous and Trustworthy Agents}

\author{\name David Peer \email david.peer@otera.ai \\
      \addr Otera \\
      Austria
      \AND
      \name Sebastian Stabinger \email sebastian.stabinger@otera.ai \\
      \addr Otera \\
      Austria
}

\makeatletter
\def\blfootnote{\gdef\@thefnmark{}\@footnotetext}
\makeatother

\def\month{MM}  
\def\year{YYYY} 
\def\openreview{\url{https://openreview.net/forum?id=XXXX}} 

\begin{document}
\raggedbottom

\maketitle
\thispagestyle{firstpage}

\begin{abstract}
Large Language Models (LLMs) have demonstrated impressive capabilities, yet their deployment in high-stakes domains is hindered by inherent limitations in trustworthiness, including hallucinations, instability, and a lack of transparency. To address these challenges, we introduce a generic neuro-symbolic approach, which we call \emph{Autonomous Trustworthy Agents (ATA)}. The core of our approach lies in decoupling tasks into two distinct phases: \textit{Offline knowledge ingestion} and \textit{online task processing}. During knowledge ingestion, an LLM translates an informal problem specification into a formal, symbolic knowledge base. This formal representation is crucial as it can be verified and refined by human experts, ensuring its correctness and alignment with domain requirements. In the subsequent task processing phase, each incoming input is encoded into the same formal language. A symbolic decision engine then utilizes this encoded input in conjunction with the formal knowledge base to derive a reliable result. Through an extensive evaluation on a complex reasoning task, we demonstrate that a concrete implementation of ATA is competitive with state-of-the-art end-to-end reasoning models in a fully automated setup while maintaining trustworthiness. Crucially, with a human-verified and corrected knowledge base, our approach significantly outperforms even larger models, while exhibiting perfect determinism, enhanced stability against input perturbations, and inherent immunity to prompt injection attacks. By generating decisions grounded in symbolic reasoning, ATA offers a practical and controllable architecture for building the next generation of transparent, auditable, and reliable autonomous agents.
\end{abstract}

\section{Introduction} \label{sec:introduction}
\blfootnote{Note: This text was corrected and improved with Gemini 2.5 Pro}

Large language models (LLMs) have demonstrated impressive capabilities across a wide spectrum of natural language processing tasks, ranging from question answering, text summarization, and machine translation to complex applications utilizing multi-agent implementations \citep{gemini-2-5,llm-coding-survey}.
They not only achieve high performance, but also allow for control via natural language prompts, making them accessible to non-experts.
This ease of interaction opens up a wide range of applications and use-cases in various industries, particularly as it empowers non-computer scientists to effectively use and control these systems. Nevertheless, especially in high-stakes industries such as insurance, healthcare, finance, and law, it is crucial that the decisions made by autonomous agents are not only accurate but also trustworthy. \citet{trustworthy-ai-overview} defines six important requirements for trustworthy AI: (R1) accuracy and stability, (R2) human agency and oversight, (R3) transparency and explainability, (R4) privacy and security, (R5) accountability, and (R6) fairness and non-discrimination. While LLMs have demonstrated impressive performance, enabling a wide range of new applications, we argue that current systems do not yet fully satisfy these trustworthy AI requirements:

\begin{figure}[t]
    \centering
    \includegraphics[width=0.7\textwidth]{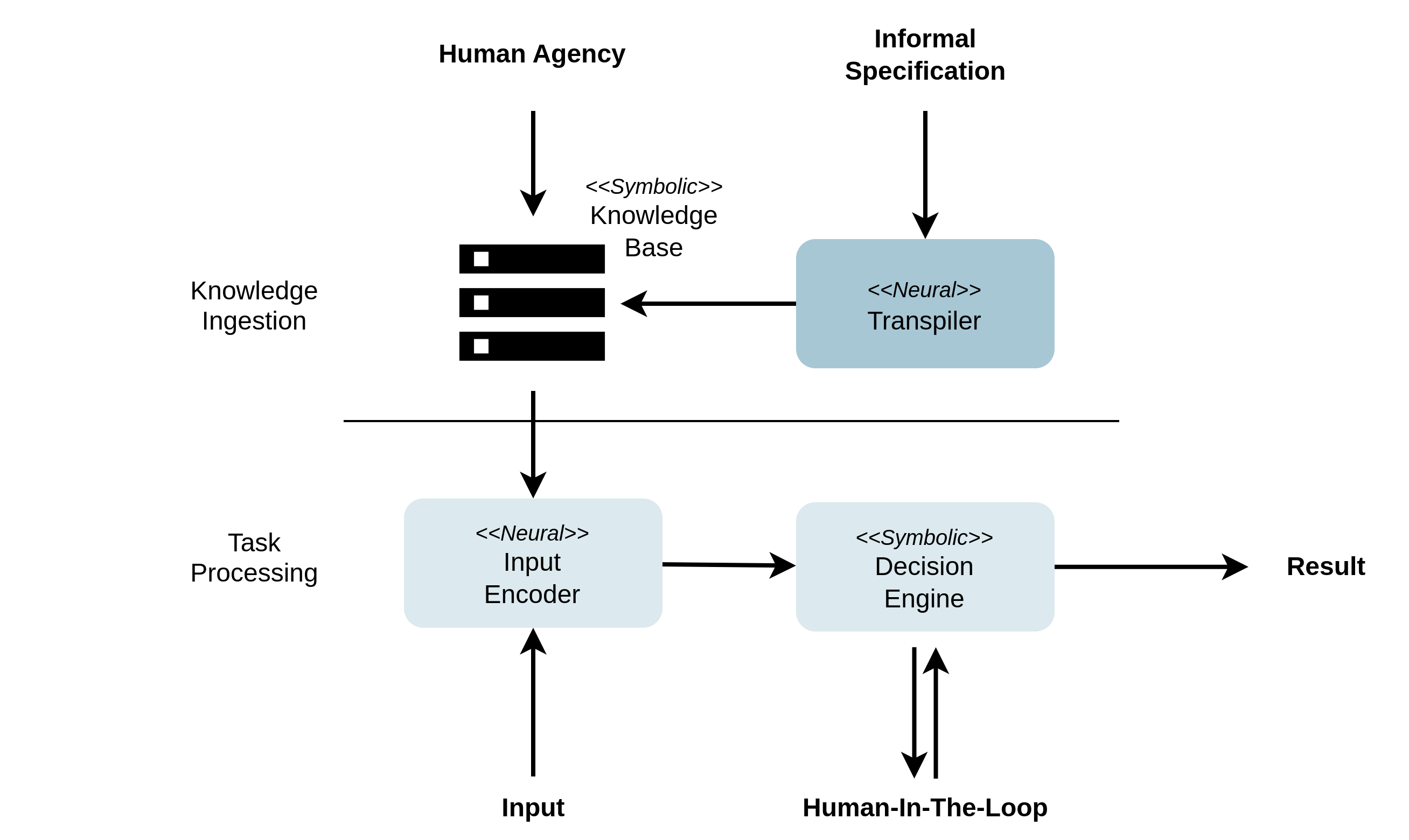}
    \caption{An overview of the proposed approach called Autonomous Trustworthy Agents (ATA).}
    \label{fig:pattern}
\end{figure}

While initial prompting of LLMs appears simple, achieving the very high accuracy required in high-stakes applications demands advanced prompt-engineering expertise and skills, as noted by \citet{prompt-engineer-requirements} (R2). This difficulty is compounded by the fact that LLMs are known to hallucinate and exhibit instability, even in purportedly deterministic setups \citep{instability-of-llms,why-llms-hallucinate} (R1). Moreover, changes to the prompt can negatively affect performance on other parts of the sample distribution, complicating proper control and necessitating a robust benchmarking and testing framework (R1, R2). It is known that reasoning LLMs contain invalid reasoning steps or hallucinated conclusions which negatively impacts the transparency and explainability of those models (R3) \citep{not-explainable-llms, reasoning-llms-are-wandering}. Furthermore, LLMs are known to be vulnerable to prompt injection attacks \citep{universal-prompt-injection} and since LLM providers typically do not offer on-premise execution, ensuring on-premise data privacy for users is often impossible (R4). Finally, LLMs exhibit biases (R6), an issue that remains an open research topic \citep{biased-llms}.

Several frameworks and approaches have been presented in the past to adress certain limitations of LLMs. For example \citet{react} presented ReAct to combine reasoning and acting in an interleaved manner in order to improve reasoning of LLMs and to make them even more generic. Instead, we introduce an approach to address the aforementioned limitations of trustworthiness, named \emph{Autonomous Trustworthy Agents (ATA)}, that combines the strengths of LLMs with the advantages of symbolic methods. Instead of querying a LLM with a single, large prompt, the task is split into two distinct building blocks: Knowledge ingestion and task processing (\autoref{fig:pattern}). The first block, knowledge ingestion, is responsible for translating an informal problem specification into a formal, symbolic knowledge base. The second block, task processing, evaluates an incoming natural language input in order to derive a reliable result w.r.t this formal knowledge base.

\emph{Knowledge ingestion} makes use of the impressive coding capabilities of LLMs \citep{llm-coding-survey} to convert an informal problem specification into a formal language. This formal representation can be an imperative programming language, such as Python or C++, but also a declarative one, such as Prolog or first-order logic. A key benefit is that the resulting formal representation can be verified and corrected by human experts ensuring that the natural language representation matches the formal representation.
This oversight allows for the removal of hallucinations and the correction of misunderstandings that can easily arise from specifications given in natural language. The \emph{task processing} step processes an unstructured, natural language input at runtime, in order to produce a desired result that complies with the knowledge base. To achieve this, the informal input is first encoded into a formal representation that complies with the knowledge base.
This step is typically a fairly well understood NLP task such as text classification, named entity recognition, or relation extraction.
Subsequently, a deterministic decision engine evaluates the formal representation of the input with respect to the formalized knowledge base. The usage of a symbolic (rather than neural) decision engine not only guarantees stability through deterministic executions, but also avoids prompt injection attacks. Additionally, a human-in-the-loop process can be triggered to review certain queries or results based on specific conditions. While ATA is a generic approach that can improve the trustworthiness significantly as shown later in this paper, it is not applicable for every task and not always straightforward to implement.

\begin{figure}[t]
    \centering
    \includegraphics[width=0.8\textwidth]{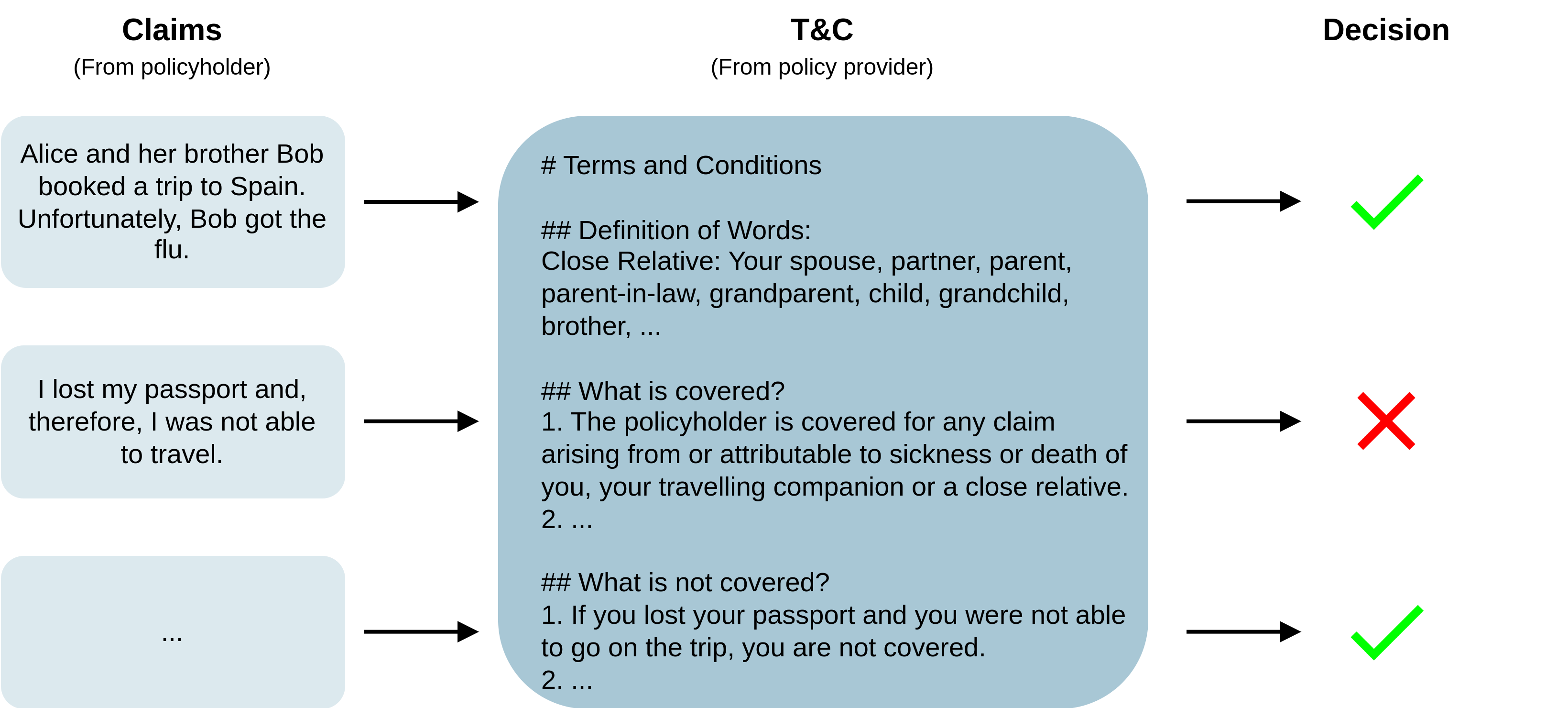}
    \caption{A graphical illustration of the claims processing task. A policyholder has signed or agreed to certain T\&Cs from a policy provider. The task is to decide if a certain claim is covered by the contract or not.}
    \label{fig:claims_processing}
\end{figure}

To demonstrate the applicability and advantages of this general approach, we also present a concrete ATA implementation for autonomous claims processing. In this use case, a policyholder files a claim under a contract governed by a certain terms and conditions (T\&Cs) policy. The T\&C document represents the informal specification while the claim is the input that needs to be processed. An autonomous agent then determines, on behalf of the policy provider, whether the claim is covered by the contract. An illustration for claims processing is given in \autoref{fig:claims_processing}. This task is really challenging, as it requires a deep understanding of the T\&Cs and the claim itself, as well as the ability to reason about the relationship between these two documents: Policies are formulated in natural language, defining rules that must be fulfilled for a claim to be covered and rules that exclude certain claims from coverage. Those rules are split over multiple sections and many pages of text. The task is therefore not only to understand a certain claim, but also to find the relevant rules in the T\&Cs and to reason about them. Rules are not only logically complex involving multiple conditions that must be satisfied, but also words used in the T\&Cs are often redefined within the context of the contract. For example, contracts might define what constitutes a relative of a policyholder.
Although T\&Cs are written in natural language, they are typically of high quality and consistency.
In contrast, claims are typically written by policyholders who lack legal expertise and may be under emotional stress due to their circumstances. This often results in claims that contain inconsistencies, missing information, or spelling errors. Consequently, implementing a robust human-in-the-loop process is essential for successful system deployment—one that can detect and handle edge cases or trigger manual review when specific conditions are met.

ATA is a valuable pattern to implement autonomous trustworthy agents for claims processing: A transpiler converts natural language T\&Cs into a declarative formal language, namely, many-sorted first-order logic. First-order logic is quite readable and therefore easy to verify and correct by human experts while still being a formal language that can be processed by automatic theorem provers. 
During task processing, the input encoder axiomizes each claim into the same formal language such that the decision engine, an automatic theorem prover, can formally prove the validity of the claim w.r.t. the T\&C. A graphical illustration and implementation of the pattern is given in \autoref{fig:idea}. While we use large reasoning models (\texttt{gemini-2.5-pro} with thinking) for the knowledge ingestion process, we demonstrate in \autoref{sec:experiments} that AT Agents that utilize small LLMs (\texttt{gemini-2.5-flash} without thinking) during task processing are very stable, with zero variance in the results, while outperforming LLMs with built-in reasoning. This can be explained by the fact that axiomization requires the LLM only for named entity recognition and relation extraction, tasks that LLMs without reasoning capabilities can solve very well. 
We also demonstrate that the controllability of the proposed system allows users to easily fix issues locally, ensuring no negative global effects on other cases.
Therefore, the performance can be significantly increased in a controlled manner, outperforming even the larger \texttt{gemini-2.5-pro} reasoning model by more than 10 percentage points, while being much more token-efficient and faster in terms of latency.
As a positive side-effect, we show that AT Agents are not longer vulnerable to prompt injection attacks, a desired property for agents that should operate autonomously.

The paper is structured as follows: In the next \autoref{sec:related_work} we discuss related work. In \autoref{sec:ata_approach} we show how to implement AT Agents for claims processing. In \autoref{sec:experiments} we present an extensive evaluation of this approach and finally conclude in \autoref{sec:conclusion_future_work}.

\section{Related Work} \label{sec:related_work}

\paragraph{LLMs and Informal Reasoning}
Large Language Models (LLMs) have already demonstrated impressive capabilities across a wide range of tasks. Subsequent work by \citet{chain-of-thought, tree-of-thought}, among others, revealed that step-by-step thinking processes can significantly enhance these results. This insight has spurred the development of several advanced reasoning models, such as those presented by \citet{gemini-2-5, llama-3, magistral, deepseek}. A survey on reinforced LLM reasoning is given by \citet{survey-reasoning-reinforced}. These models improve performance by increasing the computational budget at test-time. \citet{illusion-of-thinking} demonstrated, however, that reasoning is not universally beneficial, identifying the existence of several distinct complexity regimes. Their analysis distinguishes between low-, medium-, and high-complexity regimes, finding that reasoning provides the most significant advantage in the medium-complexity region.

\paragraph{LLMs and Formal Reasoning}
Benchmarks and a systematic evaluation of LLMs on logic tasks are provided by \citet{zebralogic}. Their findings offer critical insights into the scalability of LLM reasoning with respect to logical problems of increasing complexity.
One solution might be the combination of symbolic and neural methods, often referred to as neurosymbolic artificial intelligence. A comprehensive survey on neurosymbolic AI is provided by \citet{survey-neurosymbolic-ai}.
We particularly emphasize the integration of LLMs with formal reasoning methods to construct trustworthy systems. The concept of combining LLMs with formal methods such as propositional logic, Satisfiability Modulo Theories (SMT), and Boolean Satisfiability (SAT) solvers is a novel topic of interest. For instance, \citet{llm-smt-planning} introduced a planning framework that combines an LLM with an SMT solver to solve multi-constraint planning problems. \citet{logic-of-thought} employed LLMs to translate problems into propositional logic expressions, which can then be verified using automated theorem provers like \texttt{Prover9} \citep{prover9}. Similarly, \citet{linc} utilized first-order logic solvers to address logical reasoning challenges. In their approach, the LLM functions as a parser, translating natural language into first-order logic to derive provably correct conclusions. Perhaps the most closely related work is that of \citet{proof-of-thought} on proof reasoning. They introduced a framework and an interpreter to convert natural language statements (online) into formal logic, incorporating an error-correction loop for instances where a proof attempt fails. The output of a successful proof is subsequently cross-referenced with classical natural language reasoning to ensure a consistent conclusion or to trigger a human-in-the-loop intervention upon failure. In a subsequent publication, they demonstrated a method for estimating the confidence of a generated formal system \citep{proof-of-thought-autoformalization}.

\paragraph{LLMs and Formal Reasoning for Trustworthy AI}
Comprehensive studies on trustworthy LLMs have been conducted by \citet{trustworthy-llms-survey}, with specific examinations in high-stakes domains such as healthcare provided by \citet{study-trustworthy-llms-in-healthcare}. These studies conclude that most existing solutions are narrowly scoped and lack cross-dimensional integration, which limits their effectiveness in real-world settings with respect to the requirements for trustworthy AI outlined by \citet{trustworthy-ai-overview}. A common limitation of the aforementioned methods that combine LLMs with formal methods is their reliance on online formalization, where the LLM formalizes each new problem instance at runtime before it is passed to a theorem prover. This approach is susceptible to issues such as model hallucination or the introduction of tautologies during formalization, which can lead to erroneous conclusions. In contrast, we propose a substantially different framework for trustworthiness that decouples the formalization of the knowledge base (in our case T\&Cs) from the axiomization of a specific claim. The knowledge base formalization is performed only once for each T\&C document, with an optional step for human expert corrections in order to fix errors or to remove hallucinations. Conversely, the axiomization of a given claim is performed fully autonomously for each new instance. This separation renders the system more controllable, stable, and trustworthy, as the foundational knowledge base is static and verified before any claims are processed against it.

\section{AT Agents for Claims Processing} \label{sec:ata_approach}

\begin{figure}[t]
    \centering
    \includegraphics[width=0.7\textwidth]{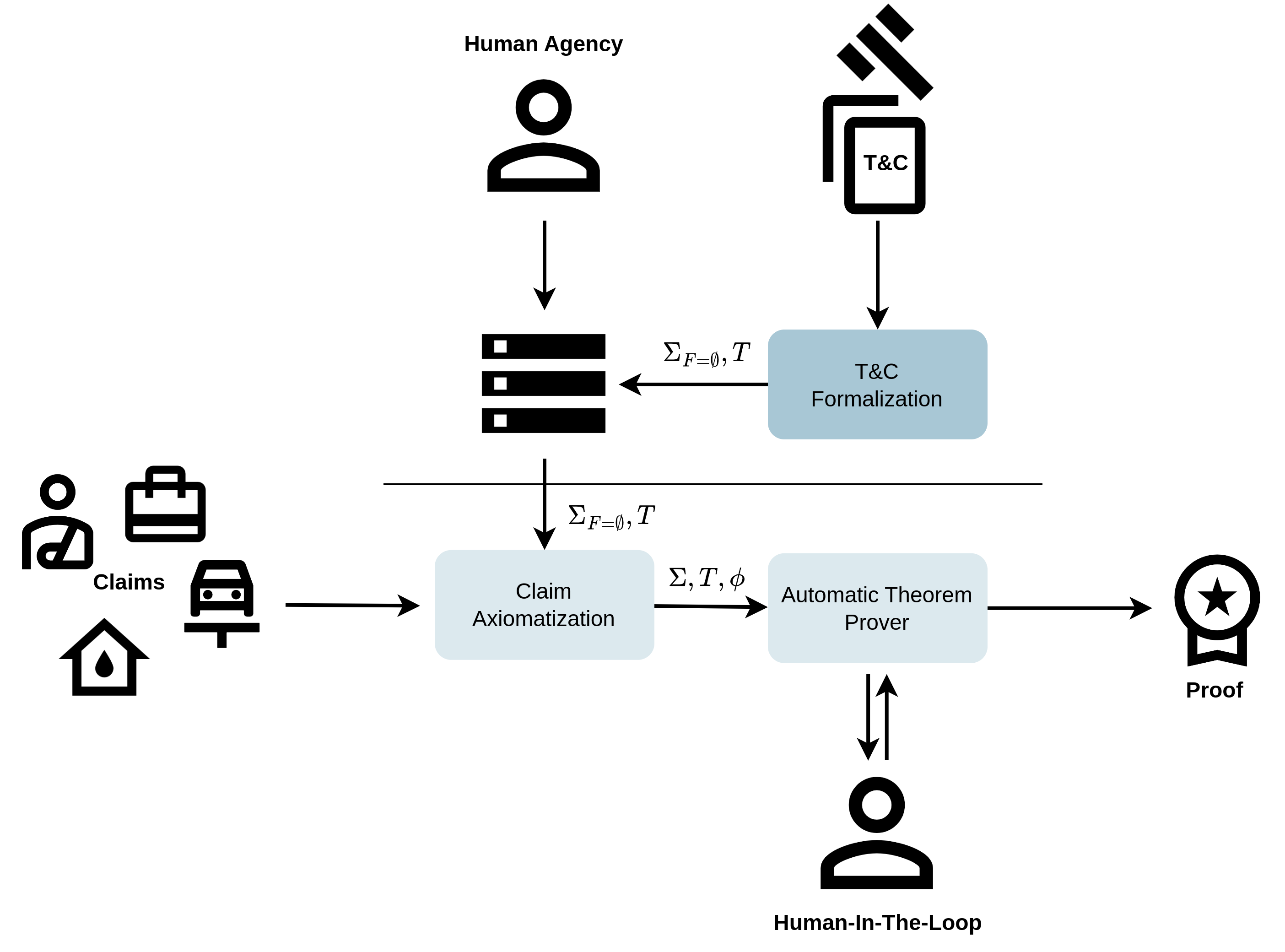}
    \caption{Concrete implementation of ATA for claims processing with knowledge ingestion and task processing: The T\&C formalization is done only once per T\&C optionally with human expert corrections, while the claim axiomization is done fully autonomously for each claim. The final decision is made by an automatic theorem prover that checks the validity of the claim w.r.t. the T\&C. The overview also shows how human experts can (1) correct the T\&C formalization and (2) review certain claims if certain conditions are met.}
    \label{fig:idea}
\end{figure}

This section demonstrates how AT Agents can be implemented for claims processing. In a typical claims processing workflow, a policyholder submits a natural language claim detailing an event for which they seek coverage. This claim must then be evaluated against the governing terms and conditions (T\&C) to determine its validity, as illustrated in \autoref{fig:claims_processing}. 

Within our proposed ATA framework, the T\&C document is first formalized into many-sorted first-order logic by a T\&C formalizer (transpiler). This one-time formalization step can optionally include verification and correction by human experts. When a new claim is submitted, it is axiomized into the same formal language to create a set of facts representing the claim (input encoding). An automatic theorem prover (decision engine) then evaluates these facts in conjunction with the formalized T\&C to determine coverage. This approach ensures that decisions are not only accurate but also transparent and explainable, as each decision corresponds to a proof that can be traced through the underlying axioms and formulas. \autoref{fig:idea} provides an overview of this implementation. 

Before the method is presented in detail, we introduce the necessary terminology and concepts from many-sorted first-order logic and satisfiability modulo theories (SMT).

\subsection{Preliminaries}
We follow the notation and definitions from \citet{combination-of-theories} for many-sorted first-order logic, but interpret a $\Sigma$-theory $T$ as a set of sentences rather than a class of structures.

\paragraph{Syntax}
A \emph{signature} $\Sigma = (\mathcal{S}, \mathcal{F}, \mathcal{P})$ is a triple consisting of a set of \emph{sorts} $\mathcal{S}$, a set of \emph{function symbols} $\mathcal{F}$, and a set of \emph{predicate symbols} $\mathcal{P}$. Each function and predicate symbol is associated with an \emph{arity}, which is a tuple of sorts from $\mathcal{S}$. Functions with an arity of a single sort are called \emph{constants}. We write \texttt{CONSTANT:Sort} to denote that the constant is of a certain sort.
We assume the standard notions of $\Sigma$-\emph{terms}, $\Sigma$-\emph{atoms}, and $\Sigma$-\emph{formulas}. A $\Sigma$-\emph{literal} is an atom or its negation. A variable is \emph{free} in a formula if it is not bound by a quantifier. A \emph{sentence} is a formula without free variables. We denote the set of variables of sort $\sigma$ occurring in a formula $\phi$ as $vars_{\sigma}(\phi)$, and the set of all variables as $vars(\phi)$.

\paragraph{Semantics}
A \emph{$\Sigma$-interpretation} $\mathcal{A}$ over a set of variables $X$ and signature $\Sigma$ is a map that interprets: Each sort $\sigma \in \mathcal{S}$ as a non-empty domain $A_{\sigma}$, each variable $x \in X$ of sort $\sigma$ as an element $x^{\mathcal{A}} \in A_{\sigma}$, each function symbol $f \in \mathcal{F}$ of arity $\sigma_1 \times \dots \times \sigma_n \rightarrow \tau$ as a function $f^{\mathcal{A}}: A_{\sigma_1} \times \dots \times A_{\sigma_n} \rightarrow A_{\tau}$, and each predicate symbol $p \in \mathcal{P}$ of arity $\sigma_1 \times \dots \times \sigma_n$ as a subset $p^{\mathcal{A}}$ of $A_{\sigma_1} \times \dots \times A_{\sigma_n}$.
The interpretation of terms and the truth-value of formulas are defined inductively based on their structure. For a term $t$, we denote its evaluation under $\mathcal{A}$ as $t^{\mathcal{A}}$. For a formula $\phi$, we denote its truth-value (\texttt{true} or \texttt{false}) as $\phi^{\mathcal{A}}$. A $\Sigma$-formula $\phi$ is \emph{satisfiable} if it evaluates to true in some $\Sigma$-interpretation over at least the variables in $vars(\phi)$. If an interpretation $\mathcal{A}$ satisfies $\phi$, we write $\mathcal{A} \models \phi$.

\paragraph{Theory} A $\Sigma$-\emph{theory} $T$ is a set of $\Sigma$-sentences. An interpretation $\mathcal{A}$ is a \emph{model} of $T$, written $\mathcal{A} \models T$, if it satisfies every sentence in $T$. A formula $\phi$ is \emph{$T$-satisfiable}, denoted as $\mathcal{A} \models_T \phi$ or simply $\mathcal{A} \models \phi$ if $T$ is clear, if there exists a model of $T$ that also satisfies $\phi$. A formula $\phi$ is \emph{$T$-valid} if it is satisfied by all models of $T$.

\subsection{Method}
The AT Agents implementation proposed in this section for claims processing defines a $\Sigma$-signature and a $\Sigma$-theory $T$ for a certain T\&C and checks the $T$-validity of a claim to decide if it is covered or not. This involves the following concrete implementations for the abstract AT components presented in \autoref{fig:pattern}:

\begin{enumerate}
  \item T\&C Formalization (Transpiler): The formalization of T\&Cs establishes the initial signature $\Sigma_{\mathcal{F}=\emptyset} = (\mathcal{S}, \emptyset, \mathcal{P})$ and the theory $T$. This foundational work is an offline process, performed only once per T\&C document, optionally allowing for human verification and corrections. 
  \item Claim Axiomization (Input Encoder): The second step is the axiomization of a claim, which extends the signature $\Sigma = (\mathcal{S}, \mathcal{F}, \mathcal{P})$ with new constants $\mathcal{F}$. All facts and relations between constants $\mathcal{F}$ w.r.t $\mathcal{P}$ are formally represented in $\phi$.
  \item Automated Theorem Proving (Decision Engine): Finally, the $T$-validity for a goal $\psi$ and claim $\phi$ is checked with an automatic theorem prover to decide if the claim is covered or not.
  The theorem prover additionally returns a minimal set of axioms and sentences that were proof-relevant or a counter example. Therefore, decisions are not only formally verified but also explainable and a human review can be triggered in case certain conditions (e.g. a certain predicate was used) are met.
\end{enumerate}
The following subsections describe all steps in more detail.

\subsubsection{T\&C Formalization (Transpiler) - Definition of $\Sigma_{\mathcal{F}=\emptyset}$ and Theory $T$}
\paragraph{$\Sigma$-sorts $\mathcal{S}$}
The first component of the signature $\Sigma_{\mathcal{F}=\emptyset}$ is the set of sorts $\mathcal{S}$. Each sort represents a fundamental class-of-interest relevant to the T\&C domain. For instance, in a travel T\&C, determining if an individual is covered necessitates a sort \texttt{Person}. For a car insurance T\&C, both the insured object and the parties involved must be tracked, requiring both a \texttt{Car} sort and a \texttt{Person} sort.

\paragraph{$\Sigma$-predicates $\mathcal{P}$}
Predicates $\mathcal{P}$ are partitioned into two distinct types: conditions and goals ($\mathcal{P} = \mathcal{P_C} \cup \mathcal{P_G}$).

\begin{enumerate}
  \item \textbf{Conditions ($\mathcal{P_C}$):} These predicates formalize the criteria that must be satisfied for a claim to be covered. For example, a T\&C rule might depend on a relative being sick. This requires the predicate \texttt{is\_sick} of type $\texttt{Person}$ and \texttt{is\_relative} of type $\texttt{Person} \times \texttt{Person}$. A crucial aspect of T\&C formalization is the redefinition of common terms. For instance, a contract may explicitly exclude cousins from the definition of a relative. To account for this, each condition predicate is defined by its symbolic name and its precise natural language definition, which is essential for accurate axiomization.
  \item \textbf{Goals ($\mathcal{P_G}$):} These predicates are used to determine the final outcome of the claim. The primary goal is represented by \texttt{is\_covered}. Depending on the T\&C, additional goal predicates may be necessary. For instance, in travel insurance, it might be essential to distinguish between coverage for trip cancellations and coverage for medical emergencies. In such cases, separate goal predicates like \texttt{is\_covered\_cancellation} and \texttt{is\_covered\_medical} would be defined.
  
\end{enumerate}

\paragraph{$\Sigma$-theory $T$}
With the signature $\Sigma$ established, the next step is to define the $\Sigma$-theory $T$. This theory comprises a set of $\Sigma$-sentences that logically encode the complex rules, conditions, and implicit properties (such as predicate symmetries or transitive relations) found within the T\&Cs. Every rule or condition in the natural language T\&Cs is systematically translated into one or more $\Sigma$-sentences using the previously defined sorts and predicates.
The core structure for these formalizations is always the following universally quantified implication:
\begin{align*}
  \forall x_1, \dots, x_n. (\mathcal{C}_1 \lor \dots \lor \mathcal{C}_k) \rightarrow G
\end{align*} 
Here, $x_1, \dots, x_n$ are variables corresponding to sorts $s_1, \dots, s_n \in \mathcal{S}$.
The antecedent is a disjunction of clauses $\mathcal{C}_1, \dots, \mathcal{C}_k$, where each clause is a conjunction of literals utilizing condition predicates $P \in \mathcal{P_C}$.
The consequent, or goal $G$, is a literal with predicate $P_G \in \mathcal{P_G}$.

The T\&C formalization, as described above, can be performed fully automatically using LLMs (in this work, we use \texttt{gemini-2.5-pro}).  Nevertheless, without human expert verification or corrections, the formalization might contain errors or hallucinations. In the experimental \autoref{sec:experiments}, we compare the performance of a fully automated T\&C formalization with a human expert-verified formalization to assess the impact of potential errors in the formalization process leading to possible future work as well.

\subsubsection{Claim Axiomization (Input Encoder) - Definition of $\Sigma, \phi$}
Given a certain claim text and a formalized T\&C ($\Sigma_{\mathcal{F}=\emptyset}, T$), the next step is to axiomize the claim by extending the signature $\Sigma_{\mathcal{F}=\emptyset}$ with $\mathcal{F}$ to $\Sigma = (\mathcal{S}, \mathcal{F}, \mathcal{P})$ and to define a set of facts or axioms $\phi$ that represent the claim. 

\paragraph{$\Sigma$-functions $\mathcal{F}$}
First, the text is checked against each sort $s \in \mathcal{S}$. Whenever an entity of sort $s$ can be identified in the claim, a corresponding $\Sigma$-constant is defined and added to $\mathcal{F}$. For example, in a travel claim "Alice is the sister of Bob" with $\mathcal{S} = \{Person\}$ we define the constants \texttt{ALICE:Person} and \texttt{BOB:Person}. This process is known as Named Entity Recognition (NER) in the NLP community. Note that NER is executed for each sort individually in order to reduce the impact of changes i.e. if a sort is changed, added or removed, there is provably no impact on any other sort.

\paragraph{Claim Axioms $\phi$}
Next, we define all facts $\phi$ that represent the natural language claim in a formal way. The axiomization $\phi$ of a claim is a conjunction of each fact $\phi_i$ that can be found in the claim such that $\phi = \bigwedge_{i=1}^m \phi_i$. An axiom $\phi_i$ is a literal with predicate $P \in \mathcal{P}$ and terms $t_1, \dots, t_n$ of sorts $s_1, \dots, s_n$ where $t_i \in \mathcal{F}$. For example, the claim "Alice is the sister of Bob" is axiomized as \texttt{is\_sister(ALICE, BOB)}. This task is known as Relation Extraction (RE) in the NLP community. Similar to the entity recognition step, relation extraction is executed for each predicate individually in order to reduce the impact of changes, and to improve the stability of the proposed approach.

Note: For both, NER as well as RE we exploit the \texttt{gemini-2.5-flash} model without reasoning as we found that small models without reasoning are sufficient for this task. 

\subsubsection{Automated Theorem Proving (Decision Engine) - Check $T$-validity}
Finally, we can prove the validity of a specific claim regarding its coverage by checking the $T$-validity of $\phi$ w.r.t a certain goal $\psi$. All goal predicates are specified in $\mathcal{P_G}$ during the T\&C formalization. For each predicate $P \in \mathcal{P_G}$, we can define a goal $\psi$ of the form $P(t_1, \dots, t_n)$ where $t_1, \dots, t_n$ are constants in $\mathcal{F}$ of sorts $s_1, \dots, s_n$ respectively. Most SMT solvers check for satisfiability. Therefore, to check the $T$-validity of a goal we need to check the $T$-unsatisfiability of the negated goal instead. E.g. to prove that \texttt{Alice} is covered, we need to check the $T$-unsatisfiability of $ \psi = \neg\texttt{is\_covered(ALICE)}$. Then we can decide if the claim is covered or not as follows: If $\psi \land \phi$ is $T$-unsat then the claim is covered. If $\psi \land \phi$ is $T$-sat then the claim is not covered and the model $\mathcal{A} \models_T \psi \land \phi$ is a counter example.

Solvers such as z3 \citep{z3} can not only find a counter-example in case of $T$-satisfiability, but in case of $T$-validity also return a so-called unsat core, which is a minimal subset of the axioms as well as sentences in $T$ that are sufficient to prove the $T$-validity of $\psi \land \phi$. Therefore, we can not only provide an answer if a claim is covered or not, but explain which constants, predicates and sentences in $T$ were used to prove the validity.
This information also enables us to trigger human review when critical predicates or rules are involved in the proof. For example, if the proof depends on a predicate indicating that a person was seriously injured, a human expert could review the case before rendering a final decision.

Finally, its worth mentioning that the satisfiability problem for first-order logic is undecidable. But since all functions in $\mathcal{F}$ are of arity 0, the Herbrand universe of $\Sigma$ is finite. Therefore, checking the $T$-satisfiability of $\psi \land \phi$ is decidable and the automatic theorem prover will always terminate with a result.

\section{Evaluation} \label{sec:experiments}

To validate the trustworthiness of AT Agents, we evaluate our system against the key properties of trustworthy AI as defined by \citet{trustworthy-ai-overview}. Depending on the nature of each property, we conduct either quantitative or qualitative evaluations.

\subsection{Setup}
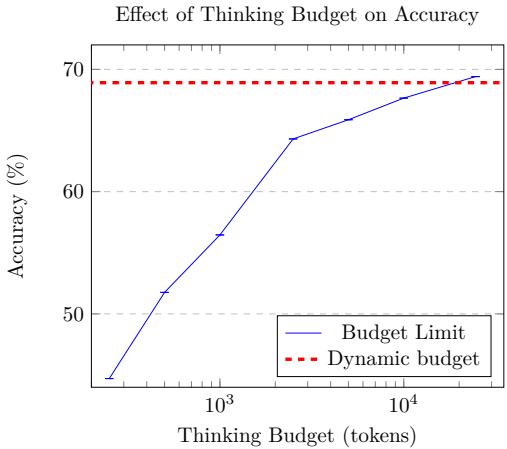
\begin{figure}[t]
    \centering
    \scalebox{.8}{
    \begin{tikzpicture}
        \begin{axis}[
            title={Effect of Thinking Budget on Accuracy},
            xlabel={Thinking Budget (tokens)},
            ylabel={Accuracy (\%)},
            xmin=200, xmax=35000,
            ymin=44, ymax=72,
            xtick={10, 100, 1000, 10000, 100000},
            ytick={50, 60, 70, 80},
            legend pos=south east,
            ymajorgrids=true,
            grid style=dashed,
            xmode=log,
        ]
        
        \addplot[
            color=blue,
            mark=no markers,
            error bars/.cd,
                y dir=both,
                y explicit,
            ]
            coordinates {
            (250, 44.71) +- (0, 0.0)
            (500, 51.76) +- (0, 0.0)
            (1000, 56.46) +- (0, 0.0)
            (2500, 64.31) +- (0, 0.0)
            (5000, 65.88) +- (0, 0.0)
            (10000, 67.65) +- (0, 0.0)
            (24576, 69.41) +- (0, 0.0)
            };

        \addplot[
            color=red,
            dashed,
            style={ultra thick},
            error bars/.cd,
                y dir=both,
                y explicit,
            ]
            coordinates {
            (1, 68.91) +- (0, 2.0)
            (100000, 68.91) +- (0, 2.0)
            };

        \legend{Budget Limit, Dynamic budget, ATA}

        \end{axis}
    \end{tikzpicture}
    }
    \caption{The effect of the thinking budget on the accuracy and standard deviation evaluated on the travel insurance dataset from $0$ to its maximum of $25\text{k}$ tokens. The blue line shows the accuracy of \texttt{gemini-2.5-flash} with reasoning and a fixed thinking budget. The red line shows the accuracy of \texttt{gemini-2.5-flash} with reasoning and a dynamic thinking budget.
    }
    \label{fig:thinking_budget_effect}
\end{figure}

\paragraph{Model:}
First of all, we selected a standard setup to compare reasoning of LLMs with the proposed AT Agents.
We use \texttt{gemini-2.5-flash} as our default model for its cost-effectiveness, speed, and strong performance on our task.
Whenever we evaluate different models, we explicitly mention it. We used the official 09/2025 deployments for gemini and version gpt-5-2025-08-07 as well as gpt-4.1-2025-04-14 for evaluating the gpt models. 
In \autoref{fig:thinking_budget_effect}, we evaluate the effect of the reasoning budget on the accuracy of claims processing. This analysis is motivated by \citet{illusion-of-thinking}, who demonstrated that reasoning can counterintuitively have a negative effect on performance in certain scenarios.
It can be concluded that reasoning is beneficial for claims processing. \texttt{gemini-2.5-flash} additionally offers a dynamic reasoning budget that automatically adjusts the reasoning budget based on the complexity of the task. We found that this dynamic budget not only matches the performance of the best fixed budget but also does not require any hyperparameter tuning. Therefore, we use it as a default in all experiments.

The proposed AT Agent for claims processing uses \texttt{gemini-2.5-flash} without thinking for input encoding (NER for defining constants $\mathcal{F}$ and RE for defining $\psi \land \phi$) to be able to compare our neuro-symbolic approach with the baseline. The transpiler (T\&C formalization) utilizes \texttt{gemini-2.5-pro} with a dynamic reasoning budget as it requires more advanced models to formalize complex T\&C rules \citep{zebralogic}.

\paragraph{Data:}
To evaluate our approach on claims processing and demonstrate its generalizability, we utilized three datasets from distinct insurance domains: travel (130 claims), electronics (240 claims), and dental (230 claims). Each dataset includes the T\&C document, ranging from 20 to 50 pages in length, along with the unstructured text of the insurance claims and its ground-truth. Due to data privacy constraints, these datasets are not publicly available at the time of writing.

\subsection{(R1) Accuracy \& Stability}
\paragraph{Accuracy:}
First we compare the accuracy of the AT Agents against the baseline LLM. More precisely, we compare our neuro-symbolic approach with built-in end-to-end reasoning capabilities of LLMs.

\begin{table}[t]
  \caption{Accuracy comparison across insurance domains.}
  \label{tbl:performance_comparison}
  \begin{center}
    \begin{tabular}{lccc}
    \toprule
    \bf Model                 & \bf Travel            & \bf Electronic        & \bf Dental   
    \\ \midrule
    \texttt{gemini-2.5-flash}   & 50.98 $\pm$ 0.55      & 60.67 $\pm$ 0.0       & 76.09 $\pm$ 0.0 \\  
    \texttt{gemini-2.5-flash} + reasoning      & 65.88 $\pm$ 1.18       & 73.13 $\pm$ 1.9       & \bf 91.5 $\pm$ 0.35 \\
    ATA              & \bf 72.94 $\pm$ 0.0   & \bf 74.00 $\pm$ 0.0   & 88.64 $\pm$ 0.0 \\
    \bottomrule
    \end{tabular}
  \end{center}
\end{table}

\autoref{tbl:performance_comparison} shows the accuracy comparison between ATA and the \texttt{gemini-2.5-flash} baseline, evaluated with and without its reasoning capabilities, across the three insurance domains.
The results lead to several key observations:
First, enabling reasoning for the baseline LLM yields a substantial performance increase of 15–20 percentage points on all datasets. This not only confirms the benefit of reasoning for this task, but also underscores the inherent complexity of determining claim coverage with respect to a given T\&C document.
Next, ATA achieves the highest accuracy on two of the three datasets and demonstrates competitive performance on the third. We wish to emphasize that our objective is not merely to outperform end-to-end reasoning LLMs in terms of accuracy, but rather to remain competitive while providing additional properties such as transparency, explainability, and stability, which are explored in the following subsections.
A key advantage of our approach is its inherent controllability. This feature allows a human expert to inspect and correct the automatic T\&C formalization, a process that, as demonstrated in a later section, can significantly improve the final accuracy.
To maintain a fair comparison with the baseline LLM---an end-to-end system that offers no intermediate checkpoints for human intervention---we evaluate the impact of human corrections separately in subsequent experiments.

\paragraph{Stability}
We evaluate ATA's robustness by assessing three distinct types of stability. First, \emph{intrinsic stability} measures the determinism of the system. To test this, we set the LLM's temperature parameter to zero and process the identical input multiple times, expecting consistent outputs. Second, \emph{extrinsic stability} assesses the system's robustness to minor, semantically neutral perturbations of the input; i.e. the approach should yield the same outcome for paraphrased inputs that do not alter the underlying meaning. Finally, \emph{LLM stability} evaluates the performance consistency when the underlying language model is substituted with a different one, thereby testing the modularity and generalizability of our approach.

The results for intrinsic stability are shown via the standard deviation in \autoref{tbl:performance_comparison}.
The end-to-end reasoning model exhibits some instability, as its complex, multi-step generation process can introduce stochasticity even at zero temperatures, as shown by \citep{non-determinism-of-llms}.
Conversely, while disabling reasoning for the baseline model increases stability, this comes at the cost of a significant drop in accuracy.
In contrast, our ATA is perfectly deterministic with competitive performance. ATA does not need built-in LLM reasoning during task processing as it confines the LLM's role to NER and RE tasks, which are sufficiently straightforward for the model to produce consistent results while the neuro-symbolic approach ensures a competitive accuracy as well. This determinism is a highly desirable property, as it allows the impact of any changes to the system to be assessed reliably.

To measure extrinsic stability, we generated three distinct, semantically equivalent paraphrases for each claim in the travel dataset using \texttt{gemini-2.5-pro}. We then evaluated the performance variation across these paraphrased inputs for both the reasoning baseline (\texttt{gemini-2.5-flash}) and ATA. The standard deviation of the accuracy metric for the baseline increased from $1.18$ to $1.47$ ($+0.29$), whereas for ATA it increased only marginally from $0.0$ to $0.15$ ($+0.15$). These results demonstrate that while both systems exhibit relative robustness to input phrasing, ATA is substantially more stable when faced with semantically equivalent variations in claim wording.

\begin{table}[t]
  \caption{The stability of the system with respect to changing the underlying LLM. Whenever possible we disabled reasoning for ATA. The superscript \textsuperscript{1} indicates that model reasoning is not available or was disabled, while the superscript \textsuperscript{2} indicates that the model has a dynamic thinking budget.}
  \label{tbl:llm_stability}
  \begin{center}
    \begin{tabular}{lccc}
    \toprule
    \bf Model               & \bf ATA & \bf ATA + Human Corr. & \bf LLM
    \\ \midrule
      \texttt{gemini-2.5-flash}      & 72.94¹   & 87.17¹         & 65.88²  \\ 
      \texttt{gemini-2.5-pro}        & 70.59²   & 83.53²         & 76.50²  \\  
      \texttt{gpt-4.1}               & 70.24¹   & 80.25¹         & 63.53¹  \\  
      \texttt{gpt-5}                 & 70.59²   & 84.71²         & 68.75² \\
    \midrule
      \bf Average           & \bf 71.09 $\pm$ 1.07 & 	\bf 83.92 $\pm$ 2.49 & \bf 68.66 $\pm$ 4.89  \\
    \bottomrule
    \end{tabular}
  \end{center}
\end{table}

Finally, we evaluated LLM stability by substituting the \texttt{gemini-2.5-flash} backbone with several alternatives: \texttt{gemini-2.5-pro}, \texttt{gpt-4.1}, and \texttt{gpt-5}. As shown in \autoref{tbl:llm_stability}, the performance of ATA remained highly stable, with a standard deviation of just $1.07$ across the different models. In contrast, the end-to-end reasoning approach was substantially less stable, exhibiting a standard deviation of $4.89$. This result demonstrates that our proposed approach is more robust to changes in the underlying LLM, highlighting its forward-compatibility with future models.
The same experiment also demonstrates that even with human corrections to the T\&C formalization (detailed in the next section), ATA maintains high stability with a standard deviation of $2.49$, compared to $4.89$ for the end-to-end reasoning LLM, while simultaneously achieving significantly higher accuracy.
Details for human corrections are explained in the next section.

\subsection{(R2) Human Agency \& Oversight}
\citet{trustworthy-ai-overview} identify human agency and oversight as fundamental requirements for trustworthy AI. Human agency ensures that users retain control over the AI system's actions, while human oversight (human-in-the-loop) provides the mechanisms for intervention and monitoring throughout the decision-making process. ATA is explicitly designed to support both principles.

\paragraph{Human Agency}
The design of ATA allows for human corrections directly after the transpilation step. This is a critical juncture where expert knowledge can be applied to verify and refine the formal representation of the specification before they are used in an automated setup. By enabling experts to review and adjust the formalization, we ensure that the system's foundational understanding of the rules aligns with human interpretations and expectations.

We demonstrate this controllability through an experiment where we manually corrected the automatically generated T\&C formalization. More precisely, we compared each natural language rule with its formalization and identified discrepancies without using any examples from our dataset. Our analysis of the initial formalization revealed three primary error categories: (1) missing clauses from the original T\&C, (2) missing logical predicates required to represent a clause, and (3) ambiguous or imprecise predicate definitions.

While the autoformalization component could undoubtedly be improved, our objective here is not to perfect this specific module. Instead, we aim to showcase the significant value of the human refinement that our architecture enables. By manually correcting the identified formalization errors, we can quantify the performance improvement directly attributable to human agency. Although this correction was performed by the authors and serves as a proof-of-concept, the experiment highlights ATA's potential to leverage expert knowledge for enhancing both reliability and accuracy.

As detailed in \autoref{tbl:performance_with_human_corrections}, these targeted manual corrections of errors 1-3 led to a significant performance improvement on the travel dataset.
As demonstrated, each category of correction contributed to a measurable increase in accuracy. This indicates that rules are independent of one another, allowing each to be corrected in isolation—a desirable property that is absent in classical prompting approaches where changes to one rule may unpredictably affect the interpretation of others.
The accuracy rose from $72.94\%$ to $87.17\%$, a substantial increase of $14.23$ percentage points. The table also provides a breakdown of this improvement, isolating the impact of correcting each category of formalization error. Crucially, after these human corrections, the ATA approach, which utilizes the efficient \texttt{gemini-2.5-flash} without reasoning for task processing, surpasses the performance of even the most advanced end-to-end reasoning LLMs, such as \texttt{gpt-5} and \texttt{gemini-2.5-pro}, by a significant margin as shown in \autoref{tbl:llm_stability}.

\begin{table}[t]
  \caption{We categorized errors and manually fixed those by comparing the natural language rule with the formalization.}
  \label{tbl:performance_with_human_corrections}
  \begin{center}
    \begin{tabular}{lc}
    \toprule
    \bf Model               & \bf Accuracy   
    \\ \midrule
      ATA without human correction              & 72.94 \\
      \qquad Added missing clauses              & \qquad + 4.71 \\
      \qquad Added missing predicates           & \qquad + 4.82 \\
      \qquad Redefined ambigious predicates     & \qquad + 4.70 \\
    \midrule
      \bf ATA with human corrections            & \bf 87.17 \\
    \bottomrule
    \end{tabular}
  \end{center}
\end{table}

\paragraph{Human Oversight}
ATA is also designed to facilitate robust human oversight through a system of configurable, rule-based triggers. These triggers can activate human-in-the-loop workflows under specific, predefined conditions (e.g. an invalid claim) to situations where the reasoning process invokes certain high-stakes rules or critical predicates. E.g. a human can be triggered whenever the predicate \texttt{is\_seriously\_injured} is proof relevant.

This mechanism enables fine-grained control over when to involve a human expert, ensuring that their attention is directed exclusively toward the most ambiguous, critical, or sensitive decisions.
Crucially, when an expert is involved, the system's inherent transparency and explainability provide the full reasoning chain that led to their involvement.
This allows the expert to quickly understand the context, make an informed decision, and permit the formal system to proceed to a consistent and verifiable final judgment as shown in the next section.

\subsection{(R3) Transparency \& Explainability}

\paragraph{Transparency}
Transparency in ATA is achieved through the formal representation of knowledge. Each rule in the T\&C is explicitly defined as a formal sentence within the theory $T$, and each fact in a claim is represented as a formal axiom in $\phi$. This explicit formalization ensures that every component of the system's knowledge base is accessible and interpretable by human experts.

\paragraph{Explainability}
The explanation generated by ATA depends on the outcome of a formal proof attempt for a given goal. For any such goal, there are two primary outcomes: it can be proven valid, or it can be found to be invalid (i.e. a counter example exists).

Given a goal is valid, a minimal unsat core can be extracted. This core contains all the axioms and rules that are required to derive that a goal is valid. This core highlights the precise set of T\&C rules and claim-specific facts that logically entail the goal. This allows an expert to efficiently trace and verify the deductive steps that led to the conclusion. Additionally, each axiom and rule can be linked into its origin in order to provide full traceability as well.

Conversely, when a goal is invalid, the prover can be used to generate counterexamples. For example, if the goal \texttt{is\_covered} is invalid, the resulting counterexample would depict a scenario consistent with all rules where the claim is not covered, thus clearly explaining the proof's failure. Standard provers such as Z3 \citep{z3} natively support both, returning a minimal unsat core in case of validity and generating a model in case of invalidity.

\subsection{(R4) Privacy \& Security}

\paragraph{Privacy}
The processing of sensitive information, such as insurance claims, demands stringent data privacy measures. The highest level of privacy is achieved when data is processed on-premise or even on-device, never leaving the owner's device. This operational constraint, however, often precludes reliance on large, cloud-based AI models and necessitates architectures that can deliver high performance using smaller, more efficient alternatives. To assess our approach's suitability for such privacy-critical and resource-constrained environments, we evaluated its performance using the most lightweight \texttt{gemini-2.5-flash-lite} model.

Remarkably, even with this lightweight backbone, the human-corrected ATA system achieved an accuracy of $76.19\%$, a result that is on-par with the performance of the much larger, state-of-the-art \texttt{gemini-2.5-pro} end-to-end model ($76.50 \%$). Results are quite behind the human-corrected ATA with \texttt{gemini-2.5-flash} ($87.17\%$), but we believe that a fine-tuned \texttt{gemini-2.5-flash-lite} model on NER and RE could close this gap further and enable high performance in secure, on-premise or on-device settings.
These experiment are left for future work.

\paragraph{Security}
A significant security advantage of our approach is its inherent immunity to prompt injection attacks. This is a direct consequence of the system's architecture, which, as illustrated in \autoref{fig:idea}, completely decouples natural language processing from the final decision-making logic. The output is generated exclusively by a formal prover that operates on structured inputs, meaning there is no \emph{natural-language bridge} from the user input to the final output that could be exploited.

The only components in the ATA approach that utilize an LLM online are for input encoding (NER and RE). However, the attack surface here is strictly limited, as (1) the output of this components is constrained to a predefined schema of predicates and sorts and (2) the output of the prover is always boolean (valid / invalid). Consequently, even if an attacker were able to manipulate the LLM's output, they could at worst alter the factual data fed to the prover. They would be unable to introduce new logical rules, modify the reasoning process itself, or inject arbitrary instructions, ensuring the integrity of the formal evaluation.

\subsection{(R5) Accountability}
As defined by \citet{trustworthy-ai-overview}, accountability involves the obligation to justify one's conduct to an authority. While establishing a complete accountability framework is a broad challenge that involves organizational policies and legal compliance, technical systems must provide the necessary foundations. Our approach contributes to this by ensuring that every decision is accompanied by a verifiable, formal proof. The high degree of transparency and explainability, derived from the unsat core or a counterexample, creates a clear audit trail. This trail allows human stakeholders to scrutinize the system's reasoning, which is a prerequisite for assigning responsibility. Although our work provides these technical enablers, the implementation of formal accountability processes remains an organizational matter beyond the scope of this paper.

\subsection{(R6) Fairness \& Non-discrimination}
It is well-established that AI systems can reflect and amplify existing societal biases, potentially leading to discriminatory outcomes for vulnerable groups \citep{biased-llms}. In our approach, this risk is still present at the axiomization step, where an LLM is used to perform NER and RE on unstructured claim data. Therefore, we want to clearly state and warn that this important challenge is not solved through the usage of our ATA approach.

\section{Conclusion} \label{sec:conclusion_future_work}
In this paper, we introduced Autonomous Trustworthy Agents (ATA), a generic neuro-symbolic approach designed to address the inherent limitations of Large Language Models (LLMs) in high-stakes domains, such as their susceptibility to hallucinations, instability, and lack of transparency. Our approach decouples task execution into two distinct phases: An offline knowledge ingestion step, where an LLM translates an informal specification into a formal, human-verifiable knowledge base, and an online task processing step, where incoming queries are handled by a deterministic symbolic decision engine. By combining the natural language understanding capabilities of LLMs with the rigor and reliability of formal methods, ATA offers a practical and controllable architecture for building the next generation of transparent, auditable, and reliable autonomous agents.

Our extensive evaluation on the complex reasoning task of autonomous claims processing demonstrates that ATA implementations are competitive with state-of-the-art end-to-end reasoning models in a fully automated setup. Crucially, we showed that by leveraging human agency to verify and correct the formalized knowledge base, our approach significantly outperforms even larger models. The resulting system exhibits perfect determinism, enhanced stability against input perturbations, and inherent immunity to prompt injection attacks. Furthermore, by grounding decisions in symbolic reasoning, ATA provides a transparent and auditable architecture where every conclusion is accompanied by a formal proof or a counterexample, enabling robust explainability and human oversight. While our approach mitigates many security and stability risks, we acknowledge that the potential for bias during the natural language understanding steps of claim axiomization remains an open challenge.

\bibliography{main}
\bibliographystyle{tmlr}

\end{document}